\documentclass{article} 
\usepackage[preprint]{colm2026_conference}

\usepackage{microtype}
\usepackage{url}
\usepackage{booktabs}
\usepackage{amsmath, amssymb, amsthm}
\usepackage{graphicx}
\usepackage{booktabs}
\usepackage{multirow}
\usepackage{xcolor}
\usepackage{hyperref}
\usepackage{url}
\usepackage{tikz}
\usepackage{pgfplots}
\usepackage{pgfplotstable}
\usepackage{natbib}
\usepackage{microtype}
\usepackage{enumitem}
\usepackage{caption}
\usepackage{subcaption}
\usepackage{algorithm}
\usepackage{algpseudocode}
\usepackage{mdframed}
\usepackage{wrapfig}
\usepackage{tcolorbox}

\usepackage{lineno}
\definecolor{darkblue}{rgb}{0, 0, 0.5}
\hypersetup{colorlinks=true, citecolor=darkblue, linkcolor=darkblue, urlcolor=darkblue}

\title{How Much Online RL is Enough?
Informative Rollouts for Offline Preference Optimization in RLVR}

\author{Richa Verma\\
TCS Research \\
Department of CSE\\
IIT Madras\\
Chennai, India \\
\texttt{richa.verma4@tcs.com} \\
\And
Balaraman Ravindran \\
Department of Data Science \& AI, \\ 
Wadhwani School of Data Science \& AI \\
IIT Madras \\
Chennai, India \\
\texttt{ravi@dsai.iitm.ac.in} \\
}

\begin{document}

\ifcolmsubmission
\linenumbers
\fi

\maketitle

\begin{abstract}
Reinforcement Learning from Verifiable Rewards (RLVR) has emerged as a 
powerful paradigm for reasoning in language models, with GRPO as its primary example. However, GRPO requires continuous online rollout 
generation, making it computationally expensive and difficult to scale. While Direct Preference 
Optimization (DPO) offers a stable and efficient offline alternative, it 
is typically expected to underperform w.r.t. online RL methods such as GRPO when trained on rollouts from a 
cold supervised fine-tuned (SFT) policy. We introduce \textbf{G2D (GRPO$\to$DPO)}, a three-stage pipeline that performs a 
short GRPO warm-up, constructs a static preference dataset, and fine-tunes a model offline with DPO. Across a set of values of the number of online steps ($K$) in GRPO on Qwen2.5-7B 
and Llama-3.1-8B, we find that \textbf{offline DPO with moderate warm-up 
matches or outperforms GRPO at substantially lower compute cost in our setting}. 
On Qwen2.5-7B, G2D at $K=150$ achieves 62.4\% on MATH-500, outperforming GRPO (51.6\%) by 10.8\% at $\sim$4$\times$ lower compute. On 
Llama-3.1-8B, G2D at $K=500$ achieves 49.4\%, surpassing GRPO in our experimental setting. We show that performance is not governed by the number of preference 
pairs, which does not vary much w.r.t. $K$, but by their 
informativeness. Moderate warm-up produces rollouts with calibrated 
uncertainty, yielding stronger contrastive signal, while excessive 
warm-up leads to overconfident policies and less informative data. Our results recast the offline-online gap in RLVR as primarily a data 
informativeness problem, and identify 
short online RL warm-up with appropriate difficulty calibration of the fine-tuning dataset as a 
compute-efficient alternative to online RL.
\end{abstract}

\section{Introduction}
\label{sec:intro}
Reinforcement Learning from Verifiable Rewards (RLVR) ~\citep{lightman2023let} has emerged as a 
powerful paradigm for improving reasoning in language models, where 
correctness is determined by a symbolic verifier rather than a learned 
reward model. Among existing approaches, Group Relative Policy Optimization 
(GRPO)~\citep{shao2024deepseekmath}, the algorithmic backbone of DeepSeek-R1~\citep{deepseekai2025deepseekr1}, has become a dominant 
method. However, GRPO is inherently online in the sense that it requires fresh rollout 
generation at every training step, making it computationally expensive and 
difficult to scale.

Direct Preference Optimization (DPO)~\citep{rafailov2023direct} provides an appealing alternative by 
eliminating online rollouts and training offline on a fixed dataset of 
preference pairs. Despite its efficiency and stability, DPO is 
generally expected to underperform in comparison to GRPO and when trained on 
rollouts from a supervised fine-tuned (SFT) policy, and this gap is 
typically attributed to distributional mismatch.

We argue that this explanation is incomplete and the primary 
bottleneck is \emph{rollout informativeness}. Effective preference 
optimization requires prompts for which both correct and incorrect 
solutions are observed in the preference dataset. Under a cold SFT~\citep{google2023gemini,openai2023gpt4} policy, most prompts likely fall 
into all-correct or all-incorrect categories, yielding weak 
contrastive signal through preference pairs. In this case, a short GRPO warm-up shifts the rollout distribution 
into a regime of calibrated uncertainty and closer to the optimal policy, where informative preference 
pairs can be constructed.

To this end, \textbf{we introduce G2D (GRPO$\to$DPO), a three-stage
pipeline that performs a short GRPO warm-up, harvests a static preference 
dataset, and trains a fresh model offline with DPO.} This allows us to harness the best of both worlds, i.e., the on-policy characteristics of online RL for mitigating data distribution shift, as well as, the scalability of offline RL. Across a set of experiments using Qwen2.5-7B~\citep{qwen2025qwen25technicalreport} and Llama-3.1-8B~\citep{dubey2024llama}, we find that offline DPO with 
moderate warm-up performs competitively with GRPO at 
substantially lower compute cost in our experimental setting.

Our results support three key findings. First, moderate warm-up is 
sufficient for offline DPO to be comparable to GRPO, while excessive 
warm-up degrades performance. Second, performance is not governed by the 
number of preference pairs, but by their 
informativeness, which peaks at intermediate warm-up. Third, the optimal 
warm-up length is model-dependent. Together, these results recast the offline-online gap in RLVR as a data quality problem under our setup. They suggest a simple and 
compute-efficient recipe: calibrate task difficulty, perform a short 
warm-up to reach an informative rollout regime, and train offline on the 
resulting preference data. \textbf{We emphasize that these results are specific to our configuration of GRPO, token budgets, and model families.}

Our contributions are as follows:
\begin{enumerate}
    \item We introduce \textbf{G2D (GRPO$\to$DPO)}, a three-stage pipeline that combines GRPO warm-up with offline DPO for RLVR. We show that offline DPO with moderate online GRPO warm-up can match or outperform GRPO at substantially lower compute cost under our setting.
    \item We identify rollout informativeness as the primary driver of performance, and quantify it via entropy and middle-band metrics.
    \item We demonstrate that the optimal warm-up length is model-dependent and closely tied to format compliance with the verifier and provide practical guidelines on difficulty calibration and rollout generation for effective offline preference optimization.
\end{enumerate}

\section{G2D: GRPO$\to$DPO}
\label{sec:method}
In this section, we provide an overview of the G2D (GRPO$\rightarrow$DPO) pipeline followed by a detailed description of the metrics used for analyzing it.
\subsection{G2D Pipeline Overview}
\begin{figure}[t]
\centering
 \includegraphics[width=0.85\linewidth, height=4.7cm]{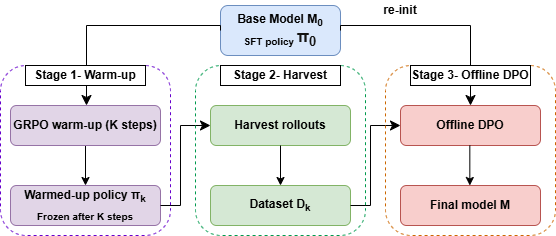}
\caption{%
\textbf{G2D pipeline.}
(\textbf{Stage 1}) Starting from a pre-trained SFT model $M_0$, we run
GRPO online for $K$ steps using LoRA. This gives us a partially warmed-up policy $\pi_k$.
(\textbf{Stage 2}) We freeze $\pi_k$ and generate rollouts for each training
prompt, scoring them with a verifiable reward to construct a static preference dataset $D_k$.
(\textbf{Stage 3}) A fresh copy of $M_0$ is then fine-tuned in an offline manner with DPO on $D_k$.}
\label{fig:pipeline}
\end{figure}

G2D is a three-stage pipeline for offline reinforcement learning~\citep{levine2020offlinereinforcementlearningtutorial} from verifiable rewards. The core idea is as follows: instead of training entirely online, we first use GRPO~\citep{shao2024deepseekmath} to perform a short online warm-up phase to move the policy into a more informative rollout space. Once this is done, we switch to learning fully offline using DPO~\citep{rafailov2023direct} on a static preference dataset harvested from the warmed-up policy in the first step. We deliberately keep rollout generation and preference optimization as two separate phases for reasons we explain below. Figure~\ref{fig:pipeline} provides an overview of the learning process. Starting from a pre-trained SFT model $M_0$, we perform $K$ steps of GRPO to obtain a partially trained policy $\pi_k$. We then freeze $\pi_k$, sample multiple rollouts per prompt, and score each rollout with a verifiable reward signal. This produces a static preference dataset $D_k$. Finally, we train a \emph{fresh} copy of $M_0$ on $D_k$ using DPO. 

\textbf{Stage 1 - Warm-up:}
We begin with a pre-trained SFT model $M_0$ parameterized by policy $\pi_0$. From this initialization point, we run GRPO for $K$ gradient steps using LoRA~\citep{hu2021loralowrankadaptationlarge} with rank 16. During this warm-up phase, we use $G=2$ rollouts per prompt and score each rollout with a binary verifiable reward $r \in \{0,1\}$ using Math-Verify.\footnote{\url{https://github.com/huggingface/Math-Verify}} Note that the purpose of this stage is not to fully solve the task online, but to shift the policy away from the cold-start regime and into a distribution slightly closer to the optimal policy $\pi^*$, where correct and incorrect rollouts co-occur more often, thereby enhancing the \emph{pairability} (see Sec. ~\ref{sec:pairability}). Setting $K=0$ is equivalent to running the standard offline DPO algorithm, in which no GRPO warm-up is performed. This gives us an effective baseline for comparison.

\textbf{Stage 2 - Harvest:}
After warm-up, we freeze the policy $\pi_k$ and use it to generate $G'=8$ rollouts per training prompt at a temperature of $\tau=1.3$. We intentionally use a higher sampling temperature than stage 1. It increases rollout diversity and makes it more likely that the same prompt yields a mix of correct and incorrect solutions. \textbf{We discuss this in detail in Sec.\ref{sec:temperature}} This mixture helps DPO learn better because preference pairs contain a meaningful contrast between a chosen and a rejected response.

Each rollout is then scored by the verifier. For a prompt to contribute to the preference dataset, it must produce at least one correct rollout and at least one incorrect rollout. When this condition is satisfied, we construct a preference pair by sampling uniformly from the two rollout classes:
\[y^+ \sim \mathcal{U}\bigl(\{y_i : r_i = 1\}\bigr),
\quad
y^- \sim \mathcal{U}\bigl(\{y_i : r_i = 0\}\bigr).\]
This ensures that the resulting dataset $D_k$ contains only those prompts where the current policy is uncertain enough to produce informative supervision.
Prompts where all $G'$ rollouts are correct (\emph{all-correct}) or all
are incorrect (\emph{all-wrong}) provide no contrastive signal and are discarded.

\textbf{Stage 3 - Offline DPO:}
We then fine-tune a fresh copy of $M_0$ on $D_k$ using the standard DPO objective \citep{rafailov2023direct} for three epochs with $\beta=0.1$. Note that the choice to re-initialize from $M_0$ rather than continue from $\pi_k$ is a critical one. It ensures that the offline DPO phase is not merely a continuation of the online GRPO policy space. Instead, it tests whether the preference data produced by the warmed-up policy in stage 2 is sufficient to train a performant final policy by itself. This design choice cleanly separates two effects that can otherwise be easily conflated:
(i) the effect of policy warm-up on the rollout distribution, and
(ii) the effect of offline learning on the final policy.
By controlling for initialization, we can cleanly attribute any changes in performance to the quality of the harvested dataset.

\subsection{Preference Pair Construction and Pairability}
\label{sec:pairability}
We define \emph{pairability} $\rho(K)$ as the fraction of training prompts that yield a valid preference pair under policy $\pi_k$. A prompt has non-zero $\rho(K)$ only if its rollout group contains at least one correct and one incorrect solution. Prompts with all-zero or all-one rewards are discarded. $\mathcal{X}_\mathrm{train}$ denotes the set of training prompts.
\[
\rho(K)
= \frac{\bigl|\{x \in \mathcal{X}_\mathrm{train}
               : 0 < \textstyle\sum_{i=1}^{G'} r_i < G'\}\bigr|}
       {|\mathcal{X}_\mathrm{train}|},
\] 
\subsection{Metrics for assessing rollout informativeness}
\label{sec:metrics}

We use the following two metrics to quantify the \emph{quality} of pairable prompts, i.e., how contrastive/informative they are.

\textbf{Middle-band fraction:}
For a pairable prompt $x$, let
\[
p(x) = \frac{1}{G'}\sum_{i=1}^{G'} r_i
\]
denote the fraction of correct rollouts. We define the middle-band fraction $\mu(K)$ as the proportion of pairable prompts for which $p(x)$ lies in the middle range. We use G' = 8 as the number of GRPO rollouts in stage 2 of G2D, therefore, the middle band or the region of highest uncertainty is given by $p(x) \in [3/8, 5/8]$. This gives us the following middle band fraction metric:
\[
\mu(K)
= \frac{\bigl|\{x : p(x) \in [3/8,\; 5/8]\}\bigr|}
       {\bigl|\{x : 0 < p(x) < 1\}\bigr|}.
\]
\textbf{These prompts are particularly valuable because the policy is genuinely uncertain on them}. This framing ensures that the correct and incorrect rollouts are both present in meaningful quantities, which leads to a stronger preference contrast and a more useful DPO signal.

\textbf{Mean rollout entropy:}
We also measure policy uncertainty by estimating the mean binary entropy of the distribution of pairable prompts:
\[
H(K)
= \frac{1}{|\{x : 0 < p(x) < 1\}|}
  \sum_{x:\, 0 < p(x) < 1} H\bigl(p(x)\bigr),
\quad
H(p) = -p\log_2 p - (1-p)\log_2(1-p).
\]
This quantity is maximized at $p=0.5$ and decreases toward zero as the rollout distribution becomes more polarized. A higher $H(K)$ indicates that the policy is uncertain over a larger fraction of pairable prompts, which is precisely the regime in which offline DPO receives the most informative (contrastive) gradients.

Empirically, we find that both $\mu(K)$ and $H(K)$ are much better predictors of downstream MATH-500~\citep{hendrycks2021measuring} performance than pairability alone (see Sec.~\ref{sec:experiments}). In our experiments, both metrics peak at moderate GRPO warm-up lengths and then decline as the policy becomes overconfident (GRPO). Therefore, ~\textbf{GRPO warm-up can shift the policy into an intermediate regime where correct and incorrect rollouts co-occur, producing higher-quality preference pairs for offline DPO}. Our experiments show that this balance is model-dependent and occurs at moderate warm-up lengths.

\section{Related Work}
\label{sec:related}

\paragraph{RLVR and GRPO for reasoning.}
In \citep{deepseekai2025deepseekr1}, the authors utilize GRPO \citep{shao2024deepseekmath} or group-relative policy optimization for
mathematical reasoning in DeepSeek-R1 to achieve state-of-the-art performance via purely online RL with verifiable rewards.
DAPO \citep{yu2025dapo} identifies entropy collapse as a key GRPO failure mode and
proposes decoupled clipping and dynamic sampling to address it. In \citet{liu2025understanding}, the authors show
that GRPO's length normalization introduces optimization bias which motivates careful hyperparameter choices during G2D warm-up stage.

\paragraph{DPO and the online-offline gap.}
DPO \citep{rafailov2023direct} reparameterizes the RLHF objective as a classification
loss on preference pairs, enabling stable offline training. However, \citet{tajwar2024preference}
and \citet{tang2024understanding} empirically show that on-policy data is critical for
preference learning mainly because offline DPO consistently underperforms online RL on reasoning tasks.
In \citet{song2024importance}, the authors provide a theoretical explanation that global coverage is
necessary for offline DPO to converge, while on-policy methods require only partial
coverage. G2D's warm-up stage directly addresses this by shifting the rollout distribution
to an uncertain region before harvesting the preference dataset for DPO.

\paragraph{Iterative and hybrid methods.}
Several works alternate between rollout generation and offline DPO updates to reap
on-policy benefits \citep{yuan2024selfrewarding, pang2024iterative, dong2024rlhf}. However, these methods remain online in spirit. The rollouts are regenerated during multiple iterations.
\textbf{G2D differs fundamentally from these methods in the sense that rollout generation occurs \emph{once} at step $K$, making
the compute requirement lower while avoiding continuous online interaction.}
The closest concurrent work, DPO-VP \citep{tu2025enhancing}, utilizes verifiable
preference pairs iteratively to match GRPO on math reasoning. On the other hand, G2D characterizes
\emph{when} a single harvest suffices and how warm-up length should be selected via diagnostic metrics such as
entropy. In \citep{wu2025ittakes}, GRPO with $G=2$ is shown to be formally
equivalent to a DPO-like contrastive loss, providing theoretical grounding for the
GRPO to DPO transition that G2D exploits.

\paragraph{Offline RL theory.}
\citet{rashidinejad2021bridging} formalize the concentrability coefficient $C^*$ for
offline RL, showing performance degrades with the density ratio between data and the
optimal policy. \citet{song2024importance} instantiate this for preference learning,
proving that coverage, not algorithm design, is the primary bottleneck. G2D's
contribution is to show empirically that \emph{rollout informativeness}, measured by
entropy and middle-band fraction, is a more actionable diagnostic for this coverage
gap than raw pairability.
\section{Experiments} 
\label{sec:experiments}
We describe the experimental setup followed by the results and their analysis below. 
\subsection{Experimental Setup}
\label{sec:setup}

\textbf{Models:}
Our primary model is Qwen2.5-7B-Instruct \citep{qwen2025qwen25technicalreport}. We also evaluate G2D on Llama-3.1-8B-Instruct \citep{dubey2024llama} to assess performance across model families. Both use different pretraining corpora and and chat templates. We fine-tune both models with LoRA (rank 16, $\alpha=32$, dropout 0.05) on a single NVIDIA A100 80GB GPU.

\textbf{Training configuration:} \textit{GRPO warm-up} phase uses two rollout groups ($G = 2$) for GRPO. Maximum completion length is set to \textbf{384 tokens for both models}. We keep the length short for this phase because the reward verifier only requires the final answer to be present, and a truncated solution that still reaches the final answer in $\backslash$\texttt{boxed\{\}} format is scored correctly. This substantially reduces the compute cost which is the dominant factor in GRPO's wall-clock time. More importantly, this introduces a trade-off: shorter completions improve efficiency but
can truncate longer reasoning chains, which may degrade reward accuracy
during GRPO training. In contrast, the harvest phase uses longer
generation budgets to preserve complete reasoning traces for preference
construction. The number of online steps in GRPO $K$ ranges over $\{0, 150, 300, 500, 700, 1000\}$. 

For \textit{rollout harvest phase}, we use a longer generation budget in comparison, i.e., \textbf{512 tokens for Qwen2.5-7B} and \textbf{896 tokens for Llama-3.1-8B}. We choose longer budget for Llama to account for its more verbose chain-of-thought style. Further, in the \textit{DPO training phase}, we initialize DPO from the base model $M_0$, not from $\pi_k$, in order to isolate the effect of data distribution from policy initialization. Next, we discuss the key results. \textbf{The appendix has all the ablations and a practical recipe for using G2D (see Appendix \ref{sec:app})}.

\textbf{Baselines:} \textbf{(i) SFT:} the base $M_0$ without any RL training.
\textbf{(ii) Offline DPO ($K=0$):} G2D with rollouts from the SFT policy $\pi_0$, equivalent to standard offline DPO on verifiable data.
\textbf{(iii) GRPO:} 1,000 steps of online GRPO with no offline phase, serving as the upper-bound compute baseline.

\textbf{We emphasize that these design choices prioritize compute efficiency
during online training and may not reflect fully optimized GRPO
configurations (e.g., longer generation budgets or larger group sizes).}
\begin{table}[t]
\centering
\small
\begin{tabular}{llcccc}
\toprule
\textbf{Model} & \textbf{Method} & \textbf{K} 
& \textbf{MATH-500} $\uparrow$ & \textbf{GSM8K} & \textbf{GPU-hrs} $\downarrow$ \\
\midrule
\multirow{5}{*}{Qwen2.5-7B}
& SFT & -- & 49.1 $\pm$ 0.4 & 90.9 & -\\
& DPO & 0 & 56.2 $\pm$ 0.8 & 91.3 & 2.2\\
& G2D & 150 & \textbf{62.4 $\pm$ 0.6} & 90.9 & 3.8 \\
& G2D & 300 & 59.2 $\pm$ 0.2 & 90.6 & 7.0 \\
& G2D & 500 & 58.5 $\pm$ 1.1 & 91.1 & 10.4\\
& G2D & 700 & 56.4 $\pm$ 0.14 & 91.1 & 14.5\\
& G2D & 1000 & 57.6 $\pm$ 1.2 & 90.4 & 20.8 \\
& GRPO (G = 2) & 1000 & 51.6 $\pm$ 0.9 & 91.5 & 26.6\\
& GRPO (G = 4) & 1000 & 53.2 $\pm$ 1.2 & 91.4 & 52.2 \\
\midrule
\multirow{5}{*}{Llama-3.1-8B}
& SFT & -- & 45.8 $\pm$ 0.1 & 84.5 & -\\
& DPO & 0 & 47.1 $\pm$ 0.7 & 84.68 & 2.6\\
& G2D & 150 & 46.4 $\pm$ 0.3 & 85.76 & 4.5\\
& G2D & 300 & 47.4 $\pm$ 0.6 & 85.29 & 7.8\\
& G2D & 500 & \textbf{49.4 $\pm$ 0.2} & 84.38 & 11.4\\
& G2D & 700 & 49.2 $\pm$ 0.4 & 84.53 & 16.8\\
& G2D & 1000 & 47.9 $\pm$ 1.6 & 84.8 & 21.8\\
& GRPO (G = 2) & 1000 & 46.1 $\pm$ 0.9 & 84.83 & 27.2\\
& GRPO (G = 4) & 1000 & 47.5 $\pm$ 0.7 & 84.9 & 54.2\\
\bottomrule
\end{tabular}
\small\caption{%
\textbf{Main results on Qwen2.5-7B-Instruct.} 
Accuracy (\%) on MATH-500 (in-distribution) and GSM8K (out-of-distribution). All accuracies are pass@1 under greedy decoding, verified by exact match
via Math-Verify, averaged over 5 random seeds.
GPU-hours are approximate on a single A100 80GB.
\textbf{Bold}: best single-run result.}
\label{tab:main}
\end{table}
\begin{figure}[t]
\centering
\includegraphics[width=0.85\linewidth]{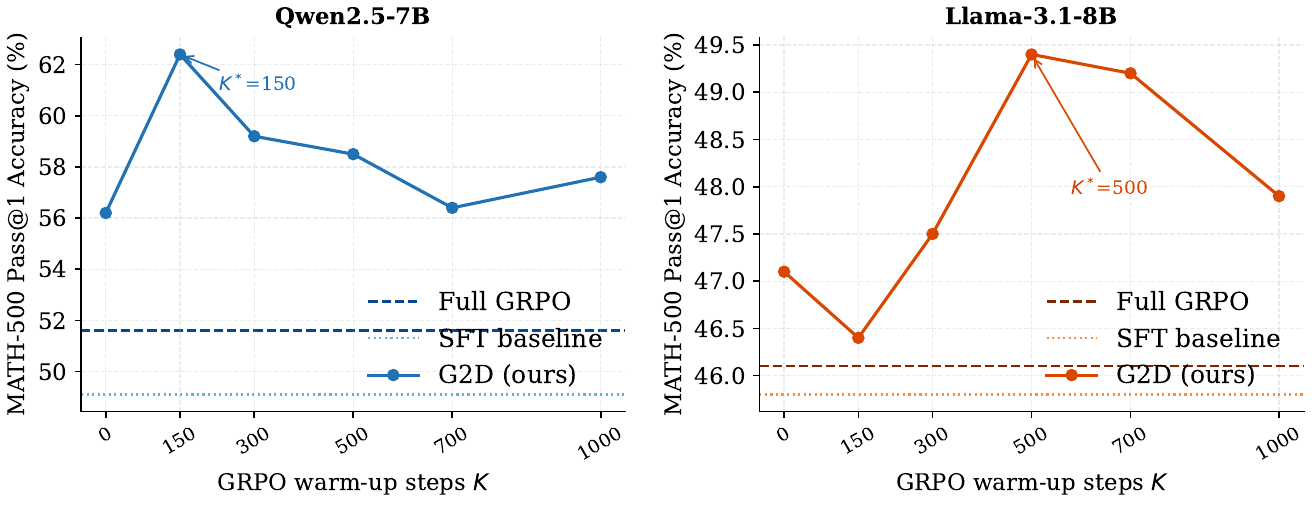}
\small\caption{%
MATH-500 accuracy vs.\ warm-up steps $K$ for Qwen2.5-7B (left) and
Llama-3.1-8B (right). Dashed lines show GRPO and SFT baselines.}
\label{fig:k_ablation}
\end{figure}
\subsection{Main Results}
\label{sec:main_results}
We report the results for both Qwen2.5-7B and Llama-3.1-8B using G2D in Table~\ref{tab:main}. All results are averaged over 5 random seeds. Figure~\ref{fig:k_ablation} shows the performance curves. \textbf{Note that all comparisons are made under similar compute budgets rather than
fully optimized hyperparameters for each method.}
\paragraph{Performance is non-monotonic and model-dependent in our experiments.}
For Qwen2.5-7B, Pass@1 accuracy peaks around $K \approx 150$-$500$
under our configuration and declines at $K \geq 700$, approaching the $K=0$ (DPO) baseline. We see the best Pass@1 accuracy with G2D at $K=150$ (62.4\%). For Llama-3.1-8B, the performance peak occurs later under our setting, at $K \approx 500$--$700$, with $K=500$ achieving 49.4\%. We discuss the mechanism behind both trends in Section~\ref{sec:pairability}.
G2D is able to improve the accuracy on MATH-500 without sacrificing it on GSM8K (out-of-distribution dataset).

\paragraph{Generalization on GSM8K is preserved across all runs.}
Despite training exclusively on MATH, all Qwen2.5-7B variants maintain
90-91\% GSM8K Pass@1 accuracy, and all Llama-3.1-8B variants maintain 84-85\% on GSM8K dataset. This confirms that neither GRPO warm-up nor offline DPO training cause catastrophic forgetting of out-of-distribution reasoning capabilities of the models used in our experiments.

\paragraph{Offline DPO with online GRPO warm-up outperforms GRPO in our setting.}
On Qwen2.5-7B, we see that G2D (G=2) across all values of $K$, i.e., the number of GRPO steps, outperforms GRPO on MATH-500 by at least 4.1\%, under the same training configuration,
while using substantially less compute. Note that the wall-clock time of GRPO is almost double that of G2D at the same number of steps $K = 1000$ and it consumes up to $4 - 5\times$ more compute. Notably, even offline DPO at $K=0$ (56.2\%) surpasses GRPO (51.6\%) by 4.6\%. This suggests that, under our training configuration, DPO is able to
extract strong learning signal from the harvested preference data,
even without continuous online rollout generation, highlighting the importance of data quality in this setting.
On Llama-3.1-8B, G2D at $K=150$ (46.4\%) outperforms GRPO (G = 2)
(46.1\%) by 0.3\% under our setting at a fraction of compute, though the advantage in this case is smaller. 

Further, to test whether the performance of GRPO is sensitive to the chosen group size (or the number of generations), we additionally train GRPO with $G=4$ rollouts per prompt and then evaluate it. This doubles the gradient signal per step at nearly twice the wall-clock time. On Qwen2.5-7B, this leads to 53.2\% on
MATH-500, a gap of only 1.4\% over GRPO with G = 2 (51.6\%). Note that this is still 9.4\%  below the best performing G2D at $K=150$ (62.4\%). Similarly, on Llama-3.1-8B tested on MATH-500, there is a marginal performance gain of 1.4\% over GRPO with G = 2. Therefore, increasing group size provides only marginal gains in our setting,
suggesting that group size alone does not explain the observed gap. \textbf{Even though GRPO typically uses $G=16$, we do not go beyond $G=4$ in our setting because the computation time would be prohibitive and no longer comparable to that of G2D.}

\paragraph{Compute efficiency.}
On Qwen2.5-7B, G2D at $K=150$ achieves 62.4\% MATH-500 at
$\approx$4 GPU-hours. This is less than one quarter of the computation time required by GRPO ($\approx$27 GPU-hours, Pass@1: 51.6\%). \textbf{Note that we interpret these results as evidence for the importance of
rollout quality under resource-constrained online training, rather than a
definitive comparison between offline and fully optimized online RL.}

\subsection{Rollout Informativeness vs.\ $K$}
\label{sec:pairability}

Intuitively, more online steps of GRPO (warm-up) should 
increase the number of usable preference pairs generated which should ideally lead to downstream gains. We test this directly and find that it does \emph{not} explain the observed gains under our setting. Table~\ref{tab:harvest} reports pairability and rollout quality metrics across $K$ for Qwen2.5-7B.

\begin{table}[h]
\centering
\small
\begin{tabular}{lccccc}
\toprule
\textbf{K} & \textbf{Pass@1} & \textbf{Pairability}
           & \textbf{N pairs} & \textbf{Entropy} $H(K)$
           & \textbf{Mid-band} $\mu(K)$ \\
\midrule
0    & 40.3\% & 47.2\% & 708 & 0.761 & 33.9\% \\
150  & 42.8\% & 46.1\% & 691 & \textbf{0.770} & \textbf{36.3\%} \\
300  & 42.8\% & 46.2\% & 693 & 0.756 & 32.8\% \\
500  & 44.1\% & 46.8\% & 702 & 0.756 & 33.5\% \\
700  & 43.1\% & 46.4\% & 696 & 0.765 & 34.9\% \\
1000 & 44.6\% & 45.3\% & 680 & 0.751 & 32.9\% \\
\midrule
\multicolumn{2}{l}{\textit{Pearson $r$ with MATH-500 Pass@1 accuracy}} & 0.021 ($\approx$ 0)& -
           & 0.253 & 0.366 \\
\bottomrule
\end{tabular}
\small\caption{%
\textbf{Harvest statistics and rollout quality metrics for various values of $K$ (Qwen2.5-7B).}
Pairability $\rho(K)$ is nearly constant and uncorrelated with
MATH-500 (Pearson $r=0.021$). Mean rollout entropy (Pearson $r=0.253$) and middle-band fraction (Pearson $r=0.366$) are substantially better predictors, both peaking at $K=150$.}
\label{tab:harvest}
\end{table}
\vspace{-5pt}
\paragraph{Pairability does not drive performance.}
The pairability rate remains nearly constant across all $K$, varying only from 45.3\% to 47.2\%, and exhibits near-zero correlation with downstream performance (Pearson $r=0.021$). This simply means that more warm-up does not help by simply increasing the number of usable/valid preference pairs. In particular, higher-performing values of $K$ do not have a higher pairability, indicating that pair count is not a key factor for accuracy.

\begin{figure}[h]
\centering
\includegraphics[width=0.8\linewidth, height = 6cm]{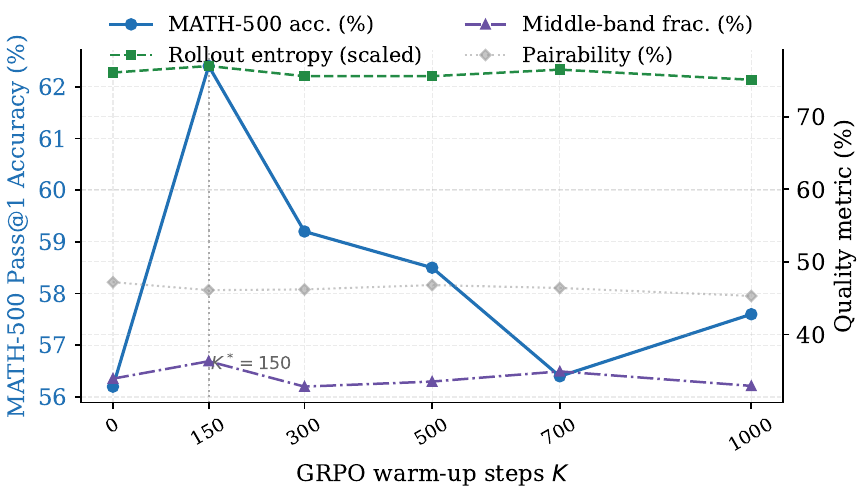}
\small\caption{%
\textbf{Rollout quality metrics vs.\ GRPO warm-up steps $K$.}
MATH-500 Pass@1 accuracy (left axis, solid) correlates with rollout entropy and
middle-band fraction (right axis, dashed/dotted), not pairability (gray,
near-flat).
All three quality metrics peak at $K=150$, explaining
why performance is non-monotonic.}
\label{fig:rollout_quality}
\end{figure}

\textbf{Pair quality drives performance.}
What changes with warm-up is the \emph{quality} of the resulting pairs. Both rollout entropy $H(K)$ and middle-band fraction $\mu(K)$ peak at intermediate $K$ and this trend closely matches that of the performance curve. Note that entropy reaches its maximum at $K=150$ (0.770) and declines thereafter. Similarly, $\mu(K)$ is highest at moderate warm-up and decreases as the policy becomes more confident. Both metrics show substantially stronger correlation with MATH-500 Pass@1 accuracy (Pearson $r=0.253$ and $r=0.366$, respectively) than pairability. Figure~\ref{fig:rollout_quality} illustrates these trends.

These effects can be explained as follows. As the value of $K$ increases, the policy $\pi_k$ improves and solves more problems correctly. Prompts that previously produced mixed rollouts (both correct and incorrect) shift into the all-correct rollouts category and are discarded during the dataset harvesting phase. The remaining pairable prompts are increasingly dominated by harder problems where the policy is consistently incorrect, resulting in low-entropy, low-contrast pairs. At moderate warm-up ($K \approx 150$), this degradation has not yet occurred, i.e., the policy is sufficiently strong to produce meaningful variation in outcomes, but still uncertain across a broad set of prompts. This regime provides us with the most informative preference pairs and, consequently, the best downstream performance. \textbf{For exemplar prompts from each stage of G2D pipeline, see Appendix ~\ref{sec:prompts}.}
 
\section{Analysis}
\label{sec:analysis}

\subsection{Why Does GRPO Underperform DPO in Our Setting?}
\label{sec:grpo_underperform}

The consistent under-performance of GRPO relative to all DPO variants in G2D
is an unusual result. We provide evidence for a likely contributing factor 
below. Note that further GRPO hyperparameter tuning may reduce this gap.

\textbf{Group size is not the explanation.}
We explore the possibility that using $G=2$ rollouts per prompt provides 
insufficient gradient signal. We test this directly by training GRPO with 
$G=4$, doubling the number of rollouts per step at nearly twice the compute cost. 
This improves the accuracy on MATH-500 only marginally, from 51.6\% to 53.0\%, and remains 
9.4 points below G2D at $K=150$ (62.4\%). Therefore, in our setting, group size does not explain the underperformance of GRPO.

\textbf{Generation length constraint during GRPO training (warm-up phase).}
In our GRPO configuration, rollouts are limited to 
\texttt{max\_completion\_length}=384 tokens to keep step time tractable given the computation resources available. 
Empirically, this truncates approximately
52\% of non-boxed Llama3.1-8B outputs mid-reasoning, and a smaller fraction for Qwen2.5-7B,  too. As a result, correct solutions with longer chains of thought are frequently assigned incorrect rewards, introducing systematic label noise during training. In contrast, preference dataset for DPO is harvested with a larger generation budget (512 tokens for Qwen and 896 tokens for Llama), preserving complete reasoning chains leading to cleaner preference supervision. Based on these observations, \textbf{we hypothesize that the generation length constraint is a key
contributing factor to the relative underperformance of the GRPO
baseline in our experiments.} A controlled ablation where GRPO can be trained on appropriate compute with longer completion lengths during the warm-up phase is a part of future work.

\subsection{Cross-Model Results: Llama-3.1-8B}
\label{sec:llama}

To assess whether our findings generalize beyond a single model family, we evaluate
G2D on Llama-3.1-8B-Instruct. Results are summarised in Table~\ref{tab:main}.
The same broad pattern holds as Qwen: G2D with moderate warm-up outperforms GRPO,
with the best result at $K=500$ (49.4\% vs.\ GRPO's 47.8\%). However, in
contrast to Qwen2.5-7B, short warm-up lengths do not yield immediate gains.

We attribute this to Llama-3.1-8B's lower base \emph{format compliance} which denotes the
fraction of rollouts that produce an answer parseable by the verifier (Math-Verify).
Specifically, a parseable answer requires the response to contain a correctly
terminated \texttt{\textbackslash boxed\{\}} expression that Math-Verify can extract
and symbolically compare against the ground truth. Empirically, at $K=0$, approximately 30\% of Llama rollouts fail this check. It does not happen necessarily because the reasoning is wrong, but
because the answer is truncated, malformatted, or something the verifier
cannot parse. This introduces systematic label noise into harvested preference pairs because genuinely correct reasoning chains are labelled as rejected, and the DPO signal is
corrupted. As warm-up length increases, format compliance improves, reducing label
noise and making preference labels more reliable. By $K=500$, rollouts are both
sufficiently accurate and consistently verifiable, enabling DPO to effectively
leverage the harvested data.

These results suggest that format compliance with respect to the verifier
is a prerequisite for effective offline preference optimization in RLVR, and that
the optimal warm-up length is model-dependent, which in turn depends on each model's baseline
capacity to produce verifiable outputs.

\subsection{Difficulty Calibration Analysis}
\label{sec:difficulty}

A key practical finding from experimenting with MATH-500 dataset is that the choice of fine-tuning difficulty levels dominates all other design choices for pairability.
Table~\ref{tab:difficulty} (see Appendix) shows per-level statistics at $K=0$. We observe that level-5 problems are too hard and that 66\% result in all-wrong rollout
groups at $K=0$, consuming a large chunk of the training budget with zero contrastive
signal in return.
Replacing level 5 with level 2 shifts the difficulty toward the ideal 50\% pass@1 regime, nearly doubling pairability (26\%$\to$47\%) without any additional compute cost. In practice, we recommend computing per-level pass@1 before
committing to a fine-tuning dataset, and picking levels where the base
model achieves roughly 30-60\% accuracy.

\section{Conclusion}
We describe G2D framework where a short GRPO warm-up followed by offline DPO can match or 
outperform online GRPO at substantially lower compute cost in our experimental
setting. The key factor is not the number of preference pairs, but their 
informativeness. Moderate warm-up produces high-quality contrastive signal, 
while excessive warm-up seems to reduce it. Our results suggest that under our setup, the offline-online gap in RLVR is primarily a data 
quality problem, governed by rollout 
informativeness, format compliance, and fine-tuning dataset difficulty calibration. Together, they provide a simple and 
compute-efficient recipe for training reasoning models that can outperform online RL under our conditions. We leave the evaluation of G2D on other popular benchmarks compatible with RLVR as future work.


\bibliography{colm2026_conference}
\bibliographystyle{colm2026_conference}

\appendix
\section{Appendix}
\label{sec:app}

\subsection{Harvest Phase: Temperature}
\label{sec:temperature}

GRPO warm-up phase uses temperature $\tau=0.8$.
We ablate the effect of higher harvest temperature on pairability during the harvest phase:

\begin{center}
\small
\begin{tabular}{lccc}
\toprule
Temperature & Pairability & All-wrong & All-correct \\
\midrule
0.8  & 23.3\% & 46.3\% & 17.0\% \\
1.3  & 47.2\% & 34.0\% & 18.8\% \\
\bottomrule
\end{tabular}
\end{center}

Higher temperature nearly doubles pairability by reducing the all-wrong
rate (46.3\%$\to$34.0\%).
This means that increasing stochasticity allows the model to explore better and to occasionally solve problems it
would otherwise fail, increasing the fraction of prompts in the mixed regime.

\subsection{Ablations}
\label{sec:ablations}

\begin{table}[h]
\centering
\small
\begin{tabular}{lccl}
\toprule
\textbf{Method} & \textbf{MATH-500} & \textbf{GSM8K} & \textbf{Note} \\
\midrule
DPO $K=0$, full (708 pairs) & 56.2\% & 91.3\% & quantity reference \\
DPO $K=0$, sub.\ (680 pairs) & 55.6\% & 91.1\% & quantity control \\
\midrule
G2D $K=150$ DPO & 62.4\% & 90.9\% & best K, baseline loss \\
G2D $K=300$ DPO & 59.2\% & 90.6\% & K ablation \\
G2D $K=300$ IPO     & 53.2\% & 80.59\% & loss fn ablation \\
\bottomrule
\end{tabular}
\caption{%
\textbf{Ablation results.}
Pair quantity has negligible effect: subsampling $D_0$ from 708 to 680
pairs costs only $-0.6\%$, confirming that pair \emph{count} does not drive performance.
IPO loss substantially underperforms DPO in this setting.}
\label{tab:ablation}
\end{table}

\textbf{Pair quantity does not drive performance.}
Subsampling $D_0$ from 708 to 680 pairs achieves
55.6\% pass@1 accuracy which is only 0.6\% below full $D_0$.
This rules out the alternative hypothesis that $K=0$ underperforms higher-$K$
variants simply because it produces fewer preference pairs.


\textbf{IPO vs.\ sigmoid DPO.}
Replacing sigmoid DPO with IPO substantially hurts performance (see
Table~\ref{tab:ablation}), indicating that the standard DPO loss is better
suited to this binary-reward preference setting.
\subsection{Difficulty Calibration: MATH-500}
\begin{table}[h]
\centering
\small
\begin{tabular}{lrrrrrr}
\toprule
\textbf{Level(s)} & \textbf{N} & \textbf{Pass@1} 
  & \textbf{Pairable} & \textbf{All-wrong} & \textbf{All-correct} \\
\midrule
3          & 421  & 61.5\% & 115\;\;(27\%) & 106\;\;(25\%) & 200\;\;(48\%) \\
4          & 447  & 46.8\% & 131\;\;(29\%) & 173\;\;(39\%) & 143\;\;(32\%) \\
5          & 632  & 20.9\% & 150\;\;(24\%) & 419\;\;(66\%) &  63\;\;(10\%) \\
\midrule
Levels 3--5 & 1500 & 40.3\% & 396\;\;(26\%) & 698\;\;(47\%) & 406\;\;(27\%) \\
Levels 2--4 & 1500 & 40.3\% & 708\;\;(47\%) & 510\;\;(34\%) & 282\;\;(19\%) \\
\bottomrule
\end{tabular}
\caption{Per-level difficulty calibration at $K=0$ (Qwen2.5-7B, $G'=8$ rollouts,
$\tau=1.3$). Level~5 contributes 66\% all-wrong rollout groups, providing zero
contrastive signal and suppressing pairability. Replacing level~5 with level~2
nearly doubles pairability (26\%$\to$47\%) at no additional compute cost.}
\label{tab:difficulty}
\end{table}

\section{Discussion}
\label{sec:discussion}
\paragraph{Practical recipe.}
Based on our findings, we recommend the following protocol for any model and math-reasoning task:
(1)~compute per-level pass@1 statistics and select difficulty levels where
the base model achieves 30-60\% accuracy;
(2)~harvest rollouts at high temperature ($\tau \geq 1.2$) with at least $G' \geq 8$
rollouts per prompt;
(3)~run a brief GRPO warm-up, monitoring rollout entropy and middle-band
fraction during training and stopping when these begin to decline;
(4)~train DPO offline on the harvested dataset, initializing from the base
model.
This pipeline consistently outperforms GRPO at substantially lower
compute cost in our experimental setting.

\paragraph{Generalization beyond math.}
The rollout informativeness perspective should apply to any RLVR setting with
binary verifiable rewards such as code generation (execution-based reward), formal proofs
(verifier-based reward), and structured NLP tasks.
The key invariant is the existence of a difficulty regime where the policy is
genuinely uncertain across rollouts, which is a prerequisite for informative
preference pairs.

\paragraph{Limitations.}
Our primary experiments are conducted on a single model family (Qwen2.5-7B)
and a single task domain (MATH reasoning).
The underperformance of GRPO in our configuration may be influenced by
our specific hyperparameter choices and further GRPO tuning (e.g., larger group
size, longer generation budget) may narrow the gap.

\textbf{Training configuration:} \textit{GRPO warm-up} phase uses \texttt{GRPOTrainer} from TRL \citep{vonWerra2020TRL} with \texttt{num\_generations}=2, \texttt{per\_device\_train\_batch\_size}=4, \texttt{gradient\_accumulation\_steps}=4, learning rate $1\times10^{-5}$, and a cosine schedule. Maximum completion length is set to \textbf{384 tokens for both
models}.

\textit{DPO training} uses TRL's \texttt{DPOTrainer} with $\beta=0.1$, learning rate
$5\times10^{-5}$, 3 training epochs and LoRA rank 16.

\section{Preference Pair Examples}
\label{sec:prompts}
This section illustrates concretely what a G2D preference pair looks like at
different warm-up lengths, using a single problem that is pairable at both $K=0$ and
$K=150$ but fully solved at $K=1000$.

\subsection{Problem and ground truth}

\begin{tcolorbox}[colback=gray!5, colframe=gray!40, title=\textbf{Problem (x)}]
A school is arranging chairs in rows for an assembly. $11$ chairs make a complete
row, and right now there are $110$ chairs total. The school wants to remove chairs
to minimize empty seats, given that $70$ students will attend. How many chairs
should be removed?
\end{tcolorbox}
\[
\text{Ground truth: } y^* = \boxed{33}
\quad \text{(verified by Math-Verify)}
\]

\subsection{$K=0$: SFT policy (2/8 correct)}

At $K=0$, only 2 of 8 rollouts produce the correct answer. The policy is
uncertain and noisy. We observe that the chosen response reaches the correct answer,
while the rejected response makes a conceptual error at a critical step (using
$\lfloor 70/11 \rfloor = 6$ rows instead of $\lceil 70/11 \rceil = 7$ rows),
arriving at $\boxed{44}$. While this pair has a valid contrastive signal, the
chosen response essentially represents a lucky correct outcome from an uncertain
policy where the reasoning is sound but the policy produces it rarely.

\begin{tcolorbox}[colback=green!4, colframe=green!40,
    title=\textbf{Chosen} $y^+$ \textnormal{(reward $r=1$)}]
\small
$\ldots$ the number of rows needed to seat 70 students is
$\lceil 70/11 \rceil = \lceil 6.36 \rceil = 7$ rows.
Total chairs needed: $7 \times 11 = 77$.
Chairs to remove: $110 - 77 = \mathbf{33}$. \quad $\boxed{33}$
\end{tcolorbox}

\begin{tcolorbox}[colback=red!4, colframe=red!30,
    title=\textbf{Rejected} $y^-$ \textnormal{(reward $r=0$)}]
\small
$\ldots$ we divide the number of students by the number of chairs per row:
$70/11 \approx 6.36$.
Since the number of rows must be a whole number, we take the integer part, which
is \textbf{6 rows}. Chairs needed: $6 \times 11 = 66$.
Chairs to remove: $110 - 66 = \mathbf{44}$. \quad $\boxed{44}$\\[2pt]
\textit{Error: uses $\lfloor 70/11 \rfloor = 6$ instead of
$\lceil 70/11 \rceil = 7$}
\end{tcolorbox}

\subsection{$K=150$: optimal warm-up (4/8 correct)}

At $K=150$, the rollouts are in the maximum-entropy regime where half of them are correct.
The chosen response follows a clean five-step solution. The rejected response
correctly computes the number of rows needed (7) and chairs required (77), but
then pivots to computing \emph{empty seats} ($110 - 70 = 40$) instead of chairs
to remove ($110 - 77 = 33$), losing the track of the objective and derailing at the final step.

\begin{tcolorbox}[colback=green!4, colframe=green!40,
    title=\textbf{Chosen} $y^+$ \textnormal{(reward $r=1$)}]
\small
(1) Rows available: $110/11 = 10$. \quad
(2) Rows needed: $\lceil 70/11 \rceil = 7$. \quad
(3) Chairs for 7 rows: $7 \times 11 = 77$. \quad
(4) Chairs to remove: $110 - 77 = \mathbf{33}$. \quad $\boxed{33}$
\end{tcolorbox}

\begin{tcolorbox}[colback=red!4, colframe=red!30,
    title=\textbf{Rejected} $y^-$ \textnormal{(reward $r=0$)}]
\small
$\ldots$ rows needed: $\lceil 70/11 \rceil = 7$ rows, total seats $= 77$. \quad
[correct so far]\\
Chairs to remove: occupied seats if 70 students are seated: $110 - 70 =
\mathbf{40}$ empty seats. [truncated, no \texttt{\textbackslash boxed\{\}}]\\[2pt]
\textit{Error: conflates ``empty seats'' with ``chairs to remove'';
computes $110 - 70$ instead of $110 - 77$, then fails to reach a final answer.}
\end{tcolorbox}

\subsection{$K=1000$: overconfident policy (8/8 correct, discarded)}

At $K=1000$, all 8 rollouts produce $\boxed{33}$ with clean reasoning. This
prompt yields no preference pair and is discarded from the dataset. As $K$
increases, an increasing fraction of training prompts enter this all-correct. This
is the key reason behind the performance decline at large $K$.

\begin{tcolorbox}[colback=gray!5, colframe=gray!40,
    title=\textbf{All rollouts correct --- prompt discarded}]
\small
All 8 rollouts: $\lceil 70/11 \rceil = 7$ rows $\Rightarrow$
$7 \times 11 = 77$ chairs $\Rightarrow$ $110 - 77 = 33$. \quad $\boxed{33}$\\[4pt]
\textit{No contrastive signal. Prompt contributes to the all-correct rate
($\uparrow$ pass@1) but provides zero DPO training pairs.}
\end{tcolorbox}

\end{document}